\documentclass[letterpaper]{article} 
\usepackage{aaai23}  
\usepackage{times}  
\usepackage{helvet}  
\usepackage{courier}  
\usepackage[hyphens]{url}  
\usepackage{graphicx} 
\usepackage{natbib}  
\usepackage{caption} 
\usepackage{algorithm}
\usepackage{algorithmic}
\usepackage{amsmath,amsfonts,bm}
\usepackage{subcaption}
\usepackage{newfloat}
\usepackage{listings}
\DeclareCaptionStyle{ruled}{labelfont=normalfont,labelsep=colon,strut=off} 
\floatstyle{ruled}
\newfloat{listing}{tb}{lst}{}
\floatname{listing}{Listing}
\def\*#1{\mathbf{#1}}
\newcommand{\+}[1]{\boldsymbol{\mathbf{#1}}}

\title{Neural Dynamic Focused Topic Model}
\author {
    Kostadin Cvejoski,\textsuperscript{\rm 1, \rm 2}
    Rams\'es J. S\'anchez, \textsuperscript{\rm 1, \rm 4}
    C\'esar Ojeda \textsuperscript{\rm 3}
}
\affiliations {
    \textsuperscript{\rm 1} Lamarr-Institute for Machine Learning and Artificial Intelligence\\
    \textsuperscript{\rm 2} Fraunhofer-Institute for Intelligent Analysis and Information Systems (IAIS)\\
    \textsuperscript{\rm 3} University of Potsdam\\
    \textsuperscript{\rm 4} BIT University of Bonn\\
    kostadin.cvejoski@iais.fraunhofer.de, sanchez@bit.uni-bonn.de, ojedamarin@uni-potsdam.de
    
}

\begin{document}

\maketitle
\renewcommand*{\thefootnote}{\fnsymbol{footnote}}
\begin{abstract}
Topic models and all their variants analyse text by learning meaningful representations through word co-occurrences. 
As pointed out by \citet{williamson2010ibp}, such models implicitly assume that the probability of a topic to be active and its proportion within each document are positively correlated.
This correlation can be strongly detrimental in the case of documents created over time, simply because recent documents are likely better described by new and hence rare topics.
In this work we leverage recent advances in neural variational inference and present an alternative neural approach to the dynamic Focused Topic Model.
Indeed, we develop a neural model for topic evolution which exploits sequences of Bernoulli random variables in order to track the appearances of topics, thereby decoupling their activities from their proportions.
We evaluate our model on three different datasets (the \textsc{UN} general debates, the collection of \textsc{NeurIPS} papers, and the \textsc{ACL} Anthology dataset) and show that it (i) outperforms state-of-the-art topic models in generalization tasks and (ii) performs comparably to them on prediction tasks, while employing roughly the same number of parameters, and converging about two times faster. Source code to reproduce our experiments is available
online.\footnote[2]{Source code: https://github.com/cvejoski/Neural-Dynamic-Focused-Topic-Model}
\end{abstract}
\renewcommand*{\thefootnote}{\arabic{footnote}}
\section{Introduction}

Probabilistic topic models, the likes of Latent Dirichlet Allocation (LDA) \cite{blei2003lda}, are generative models of word co-occurrence that analyse large document collections by learning latent representations (topics) encoding their themes.
These models represent the documents of the collection as mixtures of latent topics, and group semantically related words into single topics by means of word-pair frequency information within the collection.
Such a generic generative structure has been successfully applied to problems ranging from information retrieval, visualization and multilingual modelling to linguistic understanding in fiction and non-fiction, scientific publications and political texts (see e.g.~\citet{boyd2017applications} for a review), and keeps being extended to new domains \cite{NEURIPS2020_9f1d5659, zhao2021neural}.

Topic models implicitly assume that the documents within a given collection are exchangeable. 
Yet document collections such as magazines, academic journals, news articles and social media content not only feature trends and themes that change with time, but also employ their language differently as time evolves \cite{10.1145/2488388.2488416}. 
The exchangeability assumption along the time component is hence inappropriate in these cases and topic models have been extended to account for changes in both topic \cite{blei2006dynamic, wang2012continuous, jahnichen2018scalable} and word \cite{bamler2017dynamic, rudolph2018dynamic, dieng2019dynamic} distributions, among documents collected over long periods of time.

It is easy to imagine, however, that if one analyses the collection's content as one moves forward in time, one would find that (some of) the topics describing those documents appear, disappear or reappear with time. 
This simple intuition entails that one should not only model the time- and document-dependent topic proportions, but also \textit{the probabilities for the topics to be active}, and how such probabilities change with time. 
Previous work has already pointed out that existing topic models implicitly assume that the probability of a topic being active and its proportion within each document are positively correlated \cite{williamson2010ibp, perrone2017poisson}. 
This assumption is generally unwanted, simply because rare topics may account for a large part of the words in the few documents in which they are active.
It is particularly detrimental (for both modelling and prediction) in a dynamic setting, because recent documents are likely better described by new and hence rare topics.

Indeed, whenever the topic distribution over documents is strongly skewed, topic models tend to learn the more general topics held by the big majority of documents in the collection, rather than the rare topics contained only by fewer documents 
\cite{jagarlamudi-etal-2012-incorporating, pmlr-v32-tang14, zuo2014word}. 
Document collections that reflect evolving content typically feature skew topic distribution over its documents, with the newly added documents being well described by new, rare topics. 
Dynamic topic models that feature the topic proportion-activity coupling are then expected to perform badly, simply because these will not be able to infer the new topics characteristic of recent documents.
To properly model such recent documents one should therefore allow rarely seen topics to be active with high proportion and frequently seem topics to be active with low proportion.

In this work we seek to decouple the probability for a topic to be active from its proportion with the introduction of sequences of Bernoulli random variables, which select the active topics for a given document at a particular instant of time.
Earlier models attained such a decoupling via non-parametric priors, such as the Indian Buffet Process prior over infinite binary matrices, in both static \cite{williamson2010ibp} and dynamic \cite{perrone2017poisson} settings. 
Our construction roughly follows a similar logic, but leverages the reparametrization trick to perform  neural variational inference \cite{kingma2013auto}. 
The result is a scalable model that allows the instantaneous number of active topics per document to fluctuate, and explicitly decouples the topic proportion from its activity, thereby offering some novel layers of interpretability and transparency into the evolution of topics over time.

We introduce the Neural Dynamic Focused Topic Model (\texttt{NDF-TM}) which builds on top of Neural Variational Topic models \cite{miao2016neural} and uses Deep Kalman Filters \cite{krishnan2015deep} to model the independent dynamics of both topic proportion and topic activities.
We train and test our model on three datasets, namely the \textsc{UN} general debates, the collection of \textsc{NeurIPS} papers and the \textsc{ACL} Anthology dataset. 
Our results show via different metrics that \texttt{NDF-TM} outperforms state-of-the-art topic models in generalization tasks, and performs comparably to them on prediction tasks.
Very importantly, \texttt{NDF-TM} does this while employing roughly the same number of parameters and converging two times faster than the strongest baseline.

\section{Related Work}
\label{sec:related-work}

The \texttt{NDF-TM} model merges concepts from dynamic topic models, dynamic embeddings and neural topic models.

\textbf{Dynamic topic models}. The seminal work of \citet{blei2006dynamic} introduced the Dynamic Topic Model (DTM), which uses a state space model on the natural parameters of the distribution representing the topics, thus allowing the latter to change with time.
The DTM methodology was first extended by \citet{10.5555/3020488.3020493} to a nonparametric setting, via the correlation of Dirichlet process mixture models in time.
Later \citet{wang2012continuous} replaced the discrete state space model of DTM with a Diffusion process, thereby extending the approach to a continuous time setting.
\citet{jahnichen2018scalable} further extended DTM by introducing Gaussian process priors that allowed for a non-Markovian representation of the dynamics.
Other recent work on dynamic topic models is that of \citet{DBLP:journals/corr/abs-1805-02203}

\textbf{Dynamic embeddings}. Rather than modelling the content evolution of document collections like DTM, other works focus on modelling how word semantics change with time \cite{bamler2017dynamic, rudolph2018dynamic}. 
These works use continuous representation  of words capturing their semantics (as e.g. those of \citet{pennington2014glove}) and evolve such representation via diffusion processes. 
More recently, \citet{dieng2019dynamic} represent topics as dynamic embeddings, and model words via categorical distributions whose parameters are given by the inner product between the static word embeddings and the dynamic topic embeddings. As such, this model corresponds to the dynamic extension of \citet{dieng2020topic}.

\textbf{Neural topic models}. Another line of research leverages neural networks to improve the performance of topic models, the so-called neural topic models \cite{miao2016neural, srivastava2017autoencoding, zhang2018whai, dieng2020topic, dieng2019dynamic} which deploy neural variational inference \cite{kingma2013auto} for training. 

\textbf{Decoupling topic activity from its proportion}. \citet{williamson2010ibp} noted the implicit and undesirable correlation between topic activity and proportion assumed by standard topic models and introduced the Focused Topic Model (FTM).
FTM uses the Indian Buffet Process (IBP) to decouple across-data prevalence and within-data proportion in mixed membership models.
Later \citet{perrone2017poisson} extended FTM to a dynamic setting by using the Poisson Random Fields model from population genetics to generate dependent IBPs, which allow them to model temporal correlations in data.
Both of these models are trained using complex sampling schemes, which can make the fast and accurate inference of their model parameters difficult \cite{miao2017discovering}. 

In what follows we propose an alternative neural approach to the dynamic Focused Topic model of \citet{perrone2017poisson}, trainable via backpropagation, which learns to decouple the dynamic topic activity from its dynamic topic proportion.

\begin{figure}[t!]
     \centering
     \includegraphics[width=0.3\textwidth]{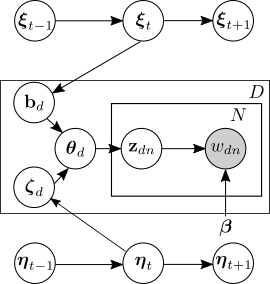}
    \caption{Graphical model representation of \texttt{NDF-TM}.}
    \label{fig:graphical_model}
\end{figure}

\section{Neural Dynamic Focused Topic Model}
\label{sec: model}

Suppose we are given an ordered collection of corpora $\mathcal{D} = \{D_1, D_2, \dots, D_T\}$, so that the $t$th corpus $D_t$ is composed of $N_t$ documents, all received within the $t$th time window. 
Let $\*W_t$ denote the Bag-of-word (BoW) representation for the whole document set within $D_t$ and let $\*w_{t,d}$ denote the BoW representation of the $d$-th document in $D_t$.

Let us now suppose that the corpora collection is described by a set of $K$ unknown topics.
We then assume there are two sequences of continuous hidden variables $\+\eta_1, \dots, \+\eta_T \in \mathbb{R}^{\mbox{\footnotesize dim}(\eta)}$ and $\+\xi_1, \dots, \+\xi_T\in \mathbb{R}^{\mbox{\footnotesize dim}(\xi)}$ which encode, respectively, how the topic proportions and the topic activities change among corpora as time evolves (i.e.~as one moves from $D_t$ to $D_{t+1}$). 
That is, $\+\eta_t$ and $\+\xi_t$ encode the \textit{global dynamics} of semantic content. 
We also assume there are two \textit{local} hidden variables, conditioned on the global ones, namely a continuous variable $\+\zeta_{t, d} \in \mathbb{R}^{K}$ which encodes the content of the $d$th document in $D_t$, in terms of the available topics, and a binary variable $\*b_{t, d} \in \{0, 1\}^{K}$ which encodes which topics are active in the document in question. 
We combine these local variables to compute the topic proportions $\+\theta_{t,d} \in [0, 1]^K$ from which the $d$th document in $D_t$ is generated. 

\subsection{Generation}

Let us denote with $\psi$ the set of parameters of our generative model.
We are first of all interested in modelling the topic activity per document at each time step, directly from the data.
One could, for example, use a $K$-dimensional mask (i.e.~a $K$-dimensional vector, whose $k$th entry is either 1 or 0 depending on whether the $k$th topic is active or inactive) for each document $d$, at each time step $t$.
To account for the variability of the data, one could also make this mask stochastic.
We thus introduce $K$ time- and document-dependent Bernoulli variables $\*b_{t, d} \in \{0, 1\}^K$ whose generation process is given by
\begin{eqnarray}
\+\xi_t     & \sim & \mathcal{N}\left(\boldsymbol{\mu}^{\xi}_{\psi}(\+\xi_{t-1}), \+\delta \, \mathbf{I}\right), \label{eq:1st-gen-model} \\
\+\pi_{t}     & = & \alpha_0 \, \mbox{Sigmoid}\left(\*W_{\xi} \, \+\xi_t + \*c_{\xi}\right),  \\
\*b_{t, d}  &\sim& \mbox{Bernoulli}(\+\pi_{t}), 
\label{eq:bernoulli-variable}
\end{eqnarray}
where $\alpha_0$ is a hyperparameter controlling the percentage of active topics,
and  $\*W_{\xi} \in \mathbb{R}^{K \times \mbox{ \footnotesize dim}(\xi)}, \*c_{\xi}\in\mathbb{R}^{K} \subset \psi$ are trainable parameters. 
Also note that, just as in Deep Kalman Filters \cite{krishnan2015deep}, $\+\xi_t$ is Markovian and evolves under a Gaussian noise with mean
$\boldsymbol{\mu}^{\xi}_{\psi}$, defined via a neural network with parameters in $\psi$, and variance $\+\delta$.  
The latter being a hyperparameter of the model. Finally, we choose $\+\xi_1 \sim \mathcal{N}(0, 1)$.

Analogously, we generate the topic proportions $\boldsymbol{\theta}_{t, d}$ as 

\begin{eqnarray}
\+\eta_t &\sim& \mathcal{N}\left(\boldsymbol{\mu}^{\eta}_\psi(\+\eta_{t-1}),\+\delta \, \mathbf{I}\right), \label{eq:transition_eta}\\
\+\zeta_{t, d} & \sim & \mathcal{N}\left(\*W_{\zeta} \, \+\eta_t + \*c_{\zeta}, \, 1\right), \\
\boldsymbol{\theta}_{t, d} & = & \frac{\*b_{t, d}\odot \exp \left( \+\zeta_{t, d}\right)}{\sum_k^K b^k_{t, d}\odot \exp \left( \zeta^k_{t, d}\right)} \label{eq:last-gen-model},
\end{eqnarray}
where $\*b_{t,d}$ is defined in \eqref{eq:bernoulli-variable} and $\odot$ labels element-wise product, $\*W_{\zeta} \in \mathbb{R}^{K \times \mbox{ \footnotesize dim}(\eta)}, \*c_{\zeta}\in\mathbb{R}^{K} \subset \psi$ are trainable, and $\boldsymbol{\mu}^{\eta}_{\psi}$ is modelled via a neural network. Here $\+\eta_t$ is also Markovian and we set $\+\eta_1 \sim \mathcal{N}(0, 1)$.
Note that the topic proportion thus defined can be \textit{sparse vectors}. That is, the model has the flexibility to completely mask some of the topics out of a given document, at a given time.

Once we have $\+\theta_{t,d}$ we generate the corpora sequence by sampling 
\begin{eqnarray}
z_{t, d, n} &\sim& \mbox{Categorical}(\boldsymbol{\theta}_{t, d}), \\
w_{t, d, n} &\sim& \mbox{Categorical}(\boldsymbol{\beta}_{z_{t, d, n}}),
\end{eqnarray}
where $z_{t, d, n}$ is the time-dependent topic assignment for $w_{t, d, n}$, which labels the $n$th word in document $d \in D_t$, and $\boldsymbol{\beta}\in\mathbb{R}^{K \times V}$ is a learnable topic distribution over words. We define the latter as
\begin{equation}
    \boldsymbol{\beta} = \mbox{softmax} (\+\alpha \otimes \+\rho),
    \label{eq:topic_distribution}
\end{equation}
with $\+\alpha \in \mathbb{R}^{K \times E}, \+\rho \in \mathbb{R}^{V \times E}$ learnable topic and word embeddings, respectively, for some embedding dimension $E$, and $\otimes$ denoting tensor product. 

\texttt{NDF-TM} is summarized in Figure \ref{fig:graphical_model}.

\begin{table*}[t!]
\centering
\begin{tabular}{l|cccccc}
\hline
            & \multicolumn{2}{c}{\textsc{UN}}        & \multicolumn{2}{c}{\textsc{NeurIPS}}    & \multicolumn{2}{c}{\textsc{ACL}} \\
Models      & PPL-DC         & P-NLL        & PPL-DC       & P-NLL        & PPL-DC    & P-NLL  \\
\hline
\texttt{DTM}*        &  2393.5 & -  &  -            & - & -                            & - \\
\texttt{DTM-REP}    &  3 012 $\pm$ 14 & 8.331 $\pm$ 0.003  &  8 107$\pm$907  & 9.5 $\pm$ 0.4 & 8 503 $\pm$ 875    & 9.7 $\pm$ 0.5 \\
\texttt{D-ETM}        & 1 748 $\pm$ 13 & \textbf{7.615 $\pm$ 0.005} &  7 746$\pm$699  & 8.983 $\pm$ 0.003 & \underline{7 805$\pm$182}            & \textbf{8.84 $\pm$ 0.02} \\
\hline
\texttt{NDF-LT-TM}    &  \underline{1 578 $\pm$ 29}  & 7.682$\pm$ 0.080 &  \underline{6 549$\pm$21}  & \underline{8.923 $\pm$ 0.002} & 7 877$\pm$213       & 8.91 $\pm$ 0.03\\
\texttt{NDF-TM}    &  \textbf{1 527 $\pm$ 36}  & \underline{7.640$\pm$ 0.004} &  \textbf{6 529$\pm$26}  & \textbf{8.901 $\pm$ 0.001} & \textbf{7 690$\pm$215}       & \underline{8.88 $\pm$ 0.03}\\
\hline
\end{tabular}
\caption{Perplexity on document completion (PPL-DC) and  predictive negative log likelihood (P-NLL). \textbf{Lower is better}.
PPL-DC is calculated by conditioning the model on the first half of the document and evaluating the perplexity on the second half of the document. 
P-NLL is estimated using equation \ref{eq: predictive_log_likelihood}.
The \texttt{DTM}* results are taken from \cite{dieng2019dynamic}.
All other results are obtained by training the models on 5 different random splits of the datasets.}
\label{tab:reults_ppl}
\end{table*}

\subsection{Inference}

The generative model above involves two independent global hidden variables $\+\xi_t, \+\eta_t$, and two local hidden variables $\+\zeta_{t,d}$ and $\*b_{t,d}$. Our task is to infer the posterior distributions of all these variables.%
\footnote{Note in passing that we do not need to perform inference of the latent topics $z_{t,d,n}$, simply because these can be integrated out (aka marginalized).} 
Denoting with $\+\Gamma_{t,d}$ the set $\{\+\xi_t, \+\eta_t, \+\zeta_{t,d}, \*b_{t,d} \}$, we approximate the true posterior distribution of the model with a variational (and structured) posterior of the form
\begin{multline}
q_{\varphi}(\+\Gamma_{t,d} | \*w_{t, d}, \*W_{1:T}) =  \\
\prod_{t}^T q_{\varphi}(\+\eta_{t} | \+\eta_{1:t-1}, \*W_{1:T}) \, q_{\varphi}(\+\xi_{t} | \+\xi_{1:t-1},\*W_{1:T}) \\
 \times \prod_{d}^{N_t}
q_{\varphi} (\+\zeta_{t, d} | \*w_{t, d}, \+\eta_t) \, q_{\varphi}(\*b_{t, d} | \*w_{t, d}, \+\xi_t),
\label{eq:posterior_distribution}
\end{multline}
where  $\*W_{1:T} = (\*W_1, \dots, \*W_T)$ is the ordered sequence of BoW representations for the corpus collection and $\varphi$ labels the variational parameters.

\textbf{Local variables}. 
The posterior distribution over the local variables $\+\zeta_{t,d}, \*b_{t,d}$ are chosen as Gaussian and Bernoulli, respectively, each parametrized by neural networks taking as input their conditional variables. 

\textbf{Global variables}. 
The posterior distribution over the dynamic global variables $\+\xi_t, \+\eta_t$ are also Gaussian, but now depend not only on the latent variables at time $t-1$, but also on the entire sequence of BoW representations $\*W_{1:T}$.
This follows directly from the graphical model in Figure \ref{fig:graphical_model}, as noted by~e.g. \citet{krishnan2015deep}. 
We shall use LSTM networks \cite{lstm} to model these dependencies.
Specifically let
\begin{equation}
q_{\varphi}(\+\xi_t|\+\xi_{t-1},\*W_{1:T}) =  \mathcal{N}(\+\mu^{\xi}_{\varphi},\+\sigma^{\xi}_{\varphi}),
\end{equation}
where $\+\mu^{\xi}_{\varphi}$, $\+\sigma^{\xi}_{\varphi}$ are neural networks which take as input the pair $(\+\xi_{t-1}, \, \*h^{\xi}_t)$, with $\*h^{\xi}_t$ a hidden representation encoding the sequence $\*W_{1:T}$. 
Similarly 
\begin{equation}
q_{\varphi}(\+\eta_t|\+\eta_{t-1},\*W_{1:T}) =  \mathcal{N}(\+\mu^{\eta}_{\varphi},\+\sigma^{\eta}_{\varphi}),
\end{equation}
where $\+\mu^{\eta}_{\varphi}, \+\sigma^{\eta}_{\varphi}$, again neural networks, take as input the pair $(\+\eta_{t-1}, \*h^{\eta}_t)$, with $\*h^{\eta}_t$ a second hidden representation also encoding $\*W_{1:T}$.
These hidden representations $\*h^i_t$, with $i=\{\xi, \eta\}$, correspond to the hidden states of LSTM networks whose update equation read
\begin{equation}
\*h^i_t = f_{\varphi}^i(\*W_t, \*h^i_{t-1}).
\label{eq:lstm_posterior}
\end{equation}

\subsection{Training Objective}

To optimize the model parameters $\{\psi, \varphi\}$ we minimize the variational lower bound on the logarithm of the marginal likelihood $p_{\psi}(w_{t, d, n}| \+\beta)$. Following standard methods \cite{bishop2006pattern}, the latter can readily be shown to be 

\begin{multline}
\small \mathcal{L}[\boldsymbol{\beta}, \psi, \varphi]  =   \sum_{t=1}^T \sum_{d=1}^{N_t}\sum_{n=1}^{N_d}\mathbb{E}_{\+\Gamma} \Big\{\log  p_{\psi}(w_{t, d, n}|\boldsymbol{\beta}, \+\Gamma) \Big\}  \\ 
\small  - \mbox{KL}\left[q_{\varphi}(\+\eta_1 | \*W_{1:T}); p(\+\eta_1)\right] - \mbox{KL}\left[q_{\varphi}(\+\xi_1 | \*W_{1:T}) ; p(\+\xi_1)\right] \\ 
\small  -\sum_{t=2}^T \mbox{KL}\left[q_{\varphi}(\+\eta_t| \+\eta_{1:t-1}, \*W_{1:T}) ; p_{\psi}(\+\eta_t| \+\eta_{t-1}) \right]   \\
\small  -\sum_{t=2}^T \mbox{KL}\left[q_{\varphi}(\+\xi_t| \+\xi_{1:t-1}, \*W_{1:T}) ; p_{\psi}(\+\xi_t| \+\xi_{t-1})\right]   \\
\small  - \sum_{t=1}^T \sum_{d=1}^{N_t} \Bigg( \mathbb{E}_{\+\eta_t} \Big\{\mbox{KL}\left[q_{\varphi}(\+\zeta_{t, d} | \*w_{t, d}, \+\eta_t); p_{\psi}(\+\zeta_{t, d} | \+\eta_{t}) \right] \Big\} \\ 
+\small  \mathbb{E}_{\+\xi_t} \Big\{ \mbox{KL}\left[q_{\varphi}(\*b_{t, d} | \*w_{t, d}, \+\xi_t); p_{\psi}(\*b_{t, d} | \+\xi_t) \right] \Big\} \Bigg),
\end{multline}
where KL labels the Kullback-Leibler divergence and $\+\beta$ is defined in Eq.~\ref{eq:topic_distribution}.

\begin{table*}[htbp]
\centering
\begin{tabular}{l|ccccccccc}
\hline
            & \multicolumn{2}{c}{\textsc{UN}}        & \multicolumn{2}{c}{\textsc{\textsc{NeurIPS}}}    & \multicolumn{2}{c}{\textsc{ACL}} \\
Models      & TC     & TD  & TC      & TD            & TC       & TD       \\
\hline
\texttt{DTM}*       & 0.1317   & 0.0799   & -   &  -     &  -  &  -   \\
\texttt{DTM-REP}    & 0.11 $\pm$ 0.30   & 0.59 $\pm$ 0.10   & -0.62$\pm$0.07   & 0.15$\pm$0.01      &  -0.82 $\pm$ 0.08  &  0.55 $\pm$ 0.02   \\
\texttt{D-ETM}       & 0.43 $\pm$ 0.20   & \underline{0.61 $\pm$0.01}   & -0.54$\pm$0.09   & 0.82 $\pm$0.01      &  \underline{-0.71$\pm$0.16}  &  0.63$\pm$0.05 \\
\hline
\texttt{NDF-LT-TM}   & \underline{0.43 $\pm$ 0.18}  &  0.56 $\pm$ 0.03   & \underline{-0.53$\pm$0.02}   &  \textbf{0.90$\pm$0.01}  &  -0.74$\pm$0.11 &  \underline{0.73$\pm$0.01} \\
\texttt{NDF-TM}   & \textbf{0.46 $\pm$ 0.20}  &  \textbf{0.63 $\pm$ 0.01}   & \textbf{-0.50$\pm$0.04}   &  \underline{0.85$\pm$0.02}  &  \textbf{-0.64$\pm$0.12 }&  \textbf{0.74$\pm$0.01} \\
\hline
\end{tabular}
\caption{Topic coherence (TC)  and Topic diversity (TD) for all models. 
\textbf{Higher is better}.
TC is calculated by taking the average pointwise  mutual information  between  two  words  drawn randomly from the same topic. 
TD is the percentage of unique words in the top 25 words of all topics. 
The \texttt{DTM}* results are taken from \cite{dieng2019dynamic}.
All other results are obtained by training the models on 5 different random splits of the datasets.
}
\label{tab:results_topic_quality}
\end{table*}

\section{Experiments}
\label{sec:experiments}

In this section we introduce our datasets and define our baselines.
Details about pre-processing and experimental setup can be found in the supplementary material, provided within the repository of our code \cite{source_code}.
Nevertheless, let us mention here that two important hyperparameters of the model are the maximum topic number $K$ and the percentage of active topics $\alpha_0$.
Both these hyperpameters are chosen via cross-validation, with $K=50$ and $\alpha_0=0.5$ given the best results\footnote{Specifically, $K$ was chosen from the set 50, 100 and 200. We found 50 to be the best value for all models, i.e. including the baselines. Similarly $\alpha_0$ was chosen from the set 0.1, 0.5, 1.0}.

\subsection{Datasets}

We evaluate our model on three datasets, namely the collection of \textsc{UN} speeches, \textsc{NeurIPS} papers and the \textsc{ACL} Anthology. 
The \textsc{UN} dataset\footnote{https://www.kaggle.com/unitednations/un-general-debates} \cite{baturo2017understanding} contains the transcription of the speeches given at the \textsc{UN} General Assembly during the period between the years 1970 and 2016. 
It consists of about 230950 documents. 
The \textsc{NeurIPS} dataset\footnote{https://www.kaggle.com/benhamner/nips-papers} contains the collection of papers published in  between the years 1987 and 2016. 
It consists of about 6562 documents. 
Finally, the \textsc{ACL} Anthology \cite{bird2008acl} contains a collection of computational linguistic and natural language processing papers published between 1973 and 2006. 
It consists of about 10514 documents. 

\subsection{Baselines}
\label{sec:baselines}

Our main aim is to study the effect of the topic proportion-activity coupling in the performance of \textit{dynamic topic models}\footnote{This means we do not consider static topic models} on data collections displaying evolving content.
To do so we compare against three models:

(1) \texttt{DTM-REP} ---  the neural extension of DTM, fitted using neural variational inference \cite{dieng2019dynamic}. 
This model uses a logistic-normal distribution, parametrized with feedforward neural networks, as posterior for the topic proportion distribution; as in \citet{miao2017discovering}.
It also uses Kalman Filters to model the topic dynamics, but parametrizes the posterior distribution over the dynamic latent variables with LSTM networks, just as in Deep Kalman Filters \cite{krishnan2015deep} (and just as \texttt{NDF-TM} too, see e.g.~Eq.~\ref{eq:lstm_posterior}).
As such, \texttt{DTM-REP} works as the dynamic extension of \citet{miao2017discovering}'s model.
It follows that the \texttt{DTM-REP} model thus defined only differs from \texttt{NDF-TM} in the way we model the topic proportions. 
Comparing our model against \texttt{DTM-REP} should therefore explicitly show the effect of lifting the topic proportion-activity coupling in dynamic neural topic models.

(2) \texttt{D-ETM} --- the Dynamic Embedded Topic Model \cite{dieng2019dynamic}, which captures the evolution of topics in such a way that both the content of topics and their proportions evolve over time.
This model adds complexity to \texttt{DTM-REP} by modelling words via categorical distributions whose parameters are given by the inner product between the static word embeddings and the dynamic topic embeddings.
In this way, \texttt{D-ETM} does not (necessarily) suffers from the topic proportion-activity coupling, for it can implicitly model their decoupling via its additional degrees of freedom. 

(3) \texttt{NDF-LT-TM} --- the Neural Dynamic Focused topic model \textit{with linear transition}.
We introduce this last baseline for the sake of ablation,
viz. to investigate the effect of the neural networks $\+\mu_{\psi}^{\xi}, \+\mu_{\psi}^{\eta}$ in Eqs.~\ref{eq:1st-gen-model} and \ref{eq:transition_eta}. \texttt{NDF-LT-TM} is defined by replacing these neural networks with the identity function.

\section{Results}
\label{sec:results}

In order to quantify the performance of our models, we first focus on two aspects, namely its prediction capabilities and its ability to generalize to unseen data. 
Later we also (qualitatively) discuss how the model actually performs the decoupling between topic activities and proportions.

(1) To test how well our models perform on a prediction task we compute the \textit{predictive negative log likelihood} (P-NLL).
Since to our knowledge the latter does not appear explicitly in the dynamic topic model literature, we briefly revisit how to estimate it in what follows.

In order to predict $N$ steps into the future we rely on the generative process of our model, albeit conditioned on the past. 
Essentially, one must generate Monte Carlo samples from the posterior distribution and propagate the latent representations ($\+\xi_t$ and $\+\eta_t$ in our model) into the future with the help of the prior transition function (Eqs. \ref{eq:1st-gen-model} and \ref{eq:transition_eta}, respectively)\footnote{Note that one is effectively performing a sequential Monte Carlo sample \cite{speekenbrink2016tutorial}, in which future steps are particles sampled from the posterior and propagated by the prior.}. 
This procedure is depicted on the conditional predictive distribution of our model
\begin{multline}
p(\*W_{T+1}|\*W_{1:T}) = \int p_{\psi}(\*W_{T+1}|\+\Gamma_{T+1})
\\ \times p_{\psi}(\+\Gamma_{T+1}|\+\Gamma_{T}) q_{\varphi}(\+\Gamma_{1:T}|\*W_{1:T}) d\+\Gamma_{1:T},
\end{multline}
where we replaced the true (intractable) posterior with the approximate posterior $q_{\varphi}(\+\Gamma_{1:T}|\*W_{1:T})$, 
and where $\+\Gamma_{t,d}$ labels the set $\{\+\xi_t, \+\eta_t, \+\zeta_{t,d}, \*b_{t,d} \}$ as before.

We can now define the predictive log likelihood as
\begin{multline}
\text{P-NLL}  =  \mathbb{E}_{p_{\psi}(\+\Gamma_{T+1}|\+\Gamma_{T})}\mathbb{E}_{q_{\varphi}(\+\Gamma_{1:T}| \*W_{1:T})} \\ \Big\{  
\log  p_{\psi}(\*W_{T+1}|\+\Gamma_{T+1}) \Big\}.
\label{eq: predictive_log_likelihood}
\end{multline}

(2) To test generalization we use three metrics, namely \textit{perplexity} (PPL) on document completion, \textit{topic coherence} (TC) and \textit{topic diversity} (TD). 
The document completion PPL is calculated on the second half of the documents in the test set, conditioned on their first half \cite{rosen2012author}. 
The TC is calculated by taking the average pointwise mutual information between two words drawn randomly from the same topic \cite{lau2014machine} and measures the interpretability of the topic.
In contrast, TD is the percentage of unique words in the top 25 words of all topics \cite{dieng2020topic}.
Note that one also often finds in the literature the \textit{topic quality} metric (TQ), 
defined as the product of TC with TD.
\subsection{Comparison with baselines} 

\begin{figure*}[t]
\centering
\includegraphics[width=.95\textwidth]{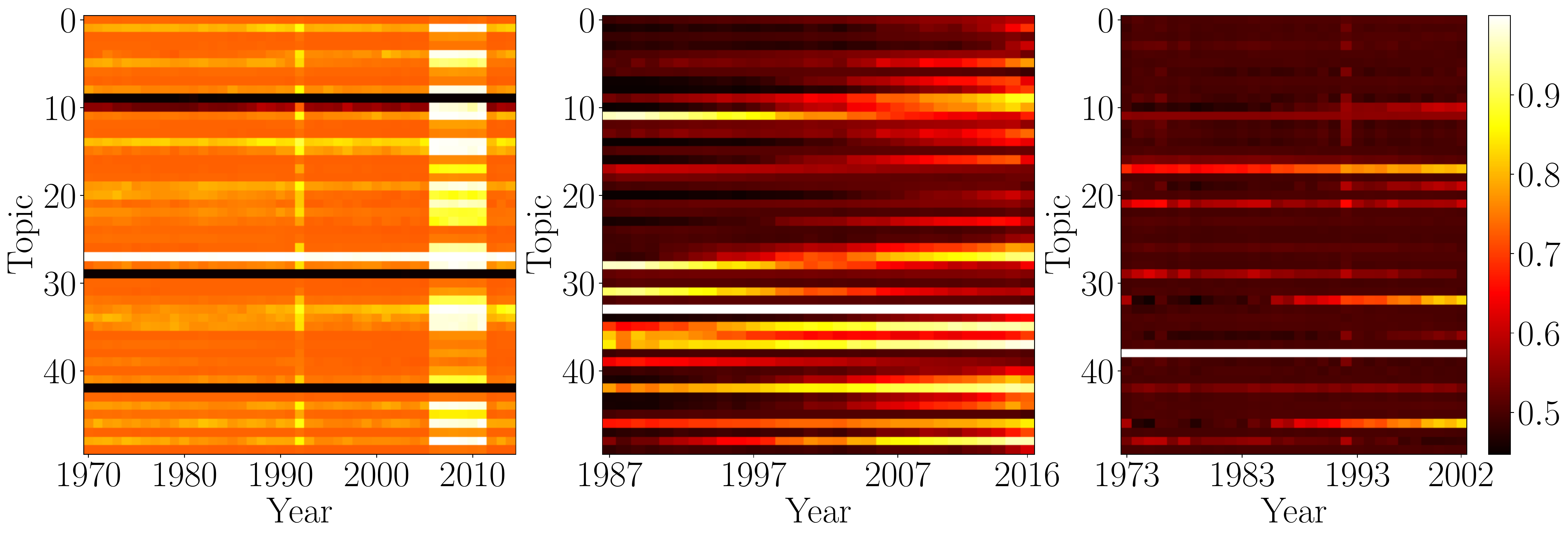}
\caption{Average time-dependent topic activity $\+b_{t,d}$ of all $K=50$ topics in \texttt{NDF-TM} for UN (left),  NIPS (middle) and \textsc{ACL} (right) datasets.}
\label{fig:b-ts}
\end{figure*}

\begin{table}[t]
\centering
\begin{tabular}{l|cccc}
\hline
Models      & 0.5         & 0.6        & 0.7    & 0.8 \\
\hline
\texttt{WF-IBP}    &    5.2	& 5.5	&6.2	& 13.8 \\
\texttt{D-ETM}      &  27.2	& 26.8	& 26.8	& 25.1  \\
\texttt{NDF-TM}     &  \textbf{35.3}	& \textbf{27.8}	& \textbf{27.8}	& \textbf{27.3} \\
\hline
\end{tabular}
\caption{Percentage (wrt. the score of the static model) of the PPL-DC difference between dynamic and static models on the \textsc{\textsc{NeurIPS}} dataset, as the percentage of held-out words was increased from 50\% to 80\%. \textbf{Higher is better}.}
\label{tab:results_perrone_modified}
\end{table}

The results on both P-NLL and PPL tasks are shown in Table \ref{tab:reults_ppl}. 
Both our models (\texttt{NDF-TM} and \texttt{NDF-LT-TM}) outperform all baselines on the completion PPL metric, on all the datasets.  
Similarly, our models outperform all baselines on both the TC and TD metrics, on all datasets, as shown in Table \ref{tab:results_topic_quality}.
These results (empirically) demonstrate that decoupling the topic activity from the topic proportion generically improves the performance of topic models on generalization tasks.
In particular, we see that adding a non-linear transformation to the prior transition functions (Eqs.~\ref{eq:1st-gen-model} and \ref{eq:transition_eta}) overall improves the model performance (i.e. compare \texttt{NDF-TM} against \texttt{NDF-LT-TM}). 

Regarding the prediction task we first notice that \texttt{NDF-TM} outperforms \texttt{DTM-REP} in all datasets.
As explained in the Baselines subsection, \texttt{DTM-REP} and \texttt{NDF-TM} only differ in the topic proportion-activity coupling, from which one can infer that lifting the coupling explicitly helps when predicting the content of future documents.
Yet \texttt{NDF-TM} only performs comparably to \texttt{D-ETM}, the strongest baseline, on this task. 
Note that \texttt{D-ETM} learns different embeddings for each topic \textit{at each time step} (i.e.~$K*T$ embeddings in total). 
One can argue that the flexibility to change the semantic content of topics as time evolves gives \texttt{D-ETM} the possibility to implicitly model rare yet relevant topics. 
In comparison, \texttt{NDF-TM} learns only $K$ topic embeddings, and has only about $\alpha_0 K$ \textit{active} embeddings (in average), at each time step.
The number of parameter for both models is about the same however,  because \texttt{NDF-TM} embeds the (fairly large) BoW vectors for the inference of its two global variables. 
Learning a single, global embedding for these BoW vectors would lower the number of needed parameters in \texttt{NDF-TM}, way below those needed in \texttt{D-ETM}, and we shall explore such an approach in the future.
\begin{figure}[t]
\begin{subfigure}{0.49\columnwidth}
  \centering
  \includegraphics[width=\columnwidth]{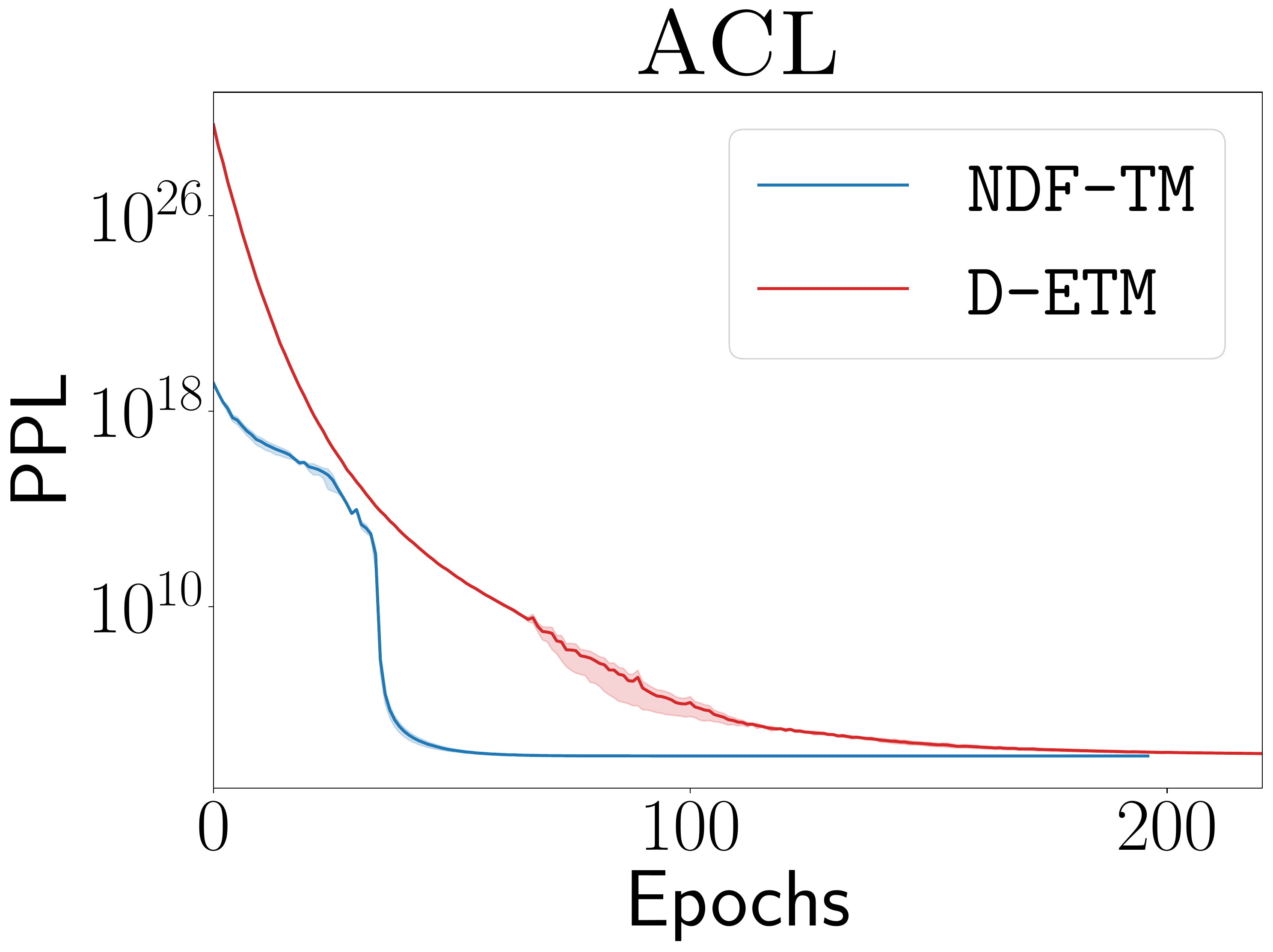} 
\end{subfigure}
\begin{subfigure}{0.49\columnwidth}
  \centering
  \includegraphics[width=\columnwidth]{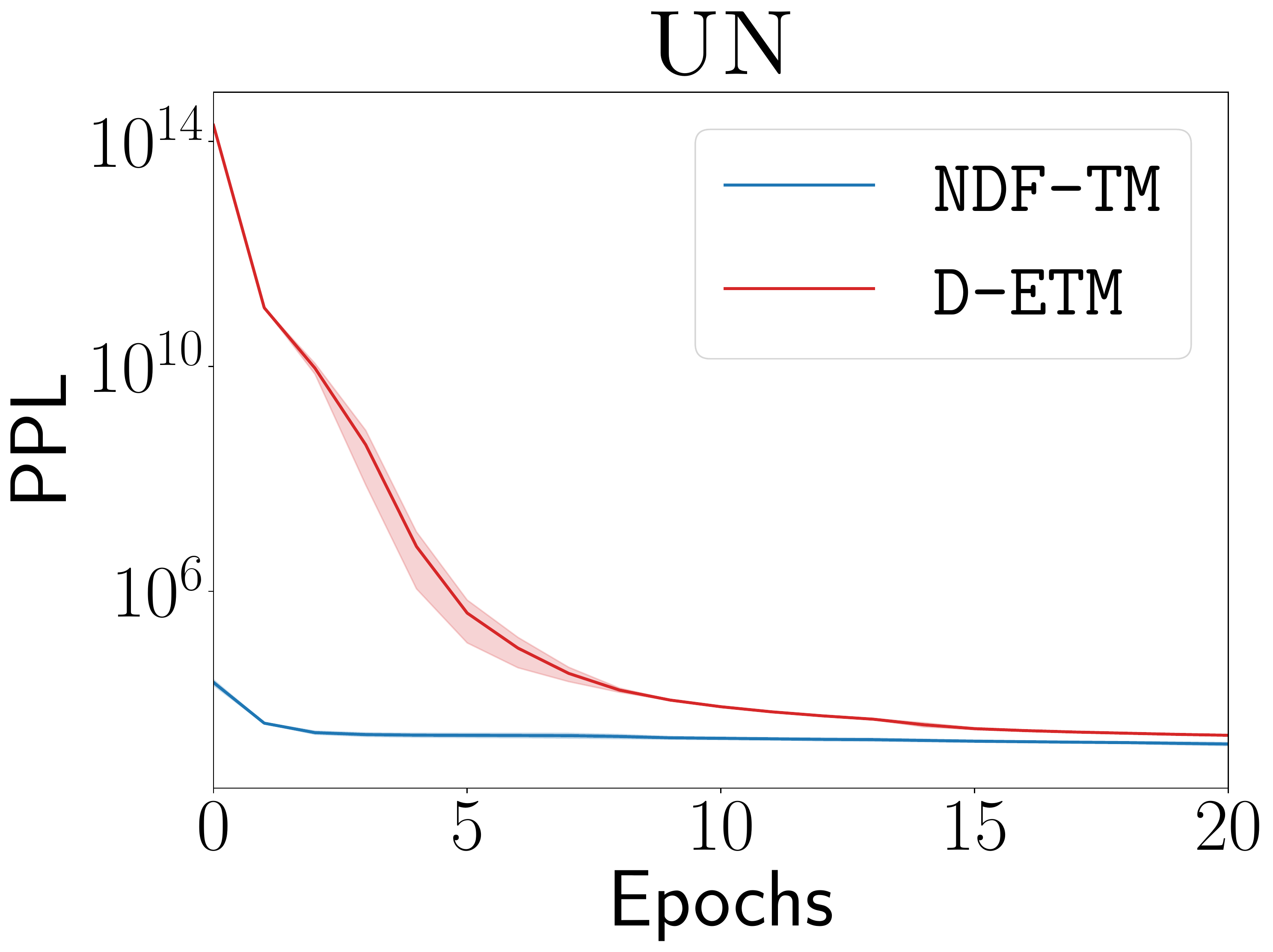}  
\end{subfigure}
\caption{Learning curves for \texttt{D-ETM} and \texttt{NDF-TM} (100 topics each) on the \textsc{ACL} and \textsc{UN} evaluation datasets. The mean and the 2x std are obtained by training the models on 5 different random splits of the data.}
\label{fig:training}
\end{figure}
Nevertheless, in practice, and as shown in Figure \ref{fig:training}, \texttt{NDF-TM} converges $\sim$2.8x faster than \texttt{D-ETM} in the \textsc{ACL} dataset (left figure). It also converges $\sim$2x faster than \texttt{D-ETM} in the \textsc{UN} dataset (right figure), and this is the worst case we have observed.
%
%
%
Thus, ultimately, \texttt{NDF-TM} is more efficient than \texttt{D-ETM}. 

We have also tried to compare against the non-parametric model of \citet{perrone2017poisson}. 
In their work they evaluated the PPL-DC on four splits of a  \textsc{NeurIPS} datasets.\footnote{Note that this dataset is different from the \textsc{NeurIPS} dataset in our main experiments. 
We only used this new one to compare against \citet{perrone2017poisson}. 
The dataset is available at  https://archive.ics.uci.edu/ml/datasets/NIPS+Conference+Papers+
1987-2015.}
The splits differ from each other on the percentage of held-out words used to define their test sets.
Intuitively, the larger the percentage of held-out words, the more a dynamic topic model has to rely on its inferred temporal representations.
The reported results seem however to be in a completely different scale from those we get (e.g. their simplest, static model yields PPL-DC values of the order of 1000, whereas our best models yield results twice as large). 
We therefore decided to compare the difference in performance between their dynamic \texttt{WF-IBP} model and their static baseline, against the difference in performance between our neural dynamic models and a static LDA model (\texttt{LDA-REP}), fitted with the reparametrization trick.
Table \ref{tab:results_perrone_modified} shows our results. 
%

\subsection{Qualitative results} 

One of our main claims is that decoupling topic activity from topic proportion helps the model better describe sequentially collected data. 
We have seen above this is indeed the case from a quantitative point of view. 
Nevertheless, one could ask whether (or how) this decoupling is effectively taking place as time evolves.
To study how the model encodes the temporal aspects of the data, we track the time evolution of both (i) the probability for topics to be active and (ii) the topic proportions. 
Figure \ref{fig:b-ts} shows the first of these.
Immediately we notice there is much more structure on the topic activities in both the \textsc{NeurIPS} and \textsc{ACL} datasets, as compared to the \textsc{UN} dataset.
We can understand these findings by arguing (\textit{a posteriori}) that \textsc{NeurIPS} and \textsc{ACL} feature more emergent and volatile  topics (wrt. their activity) as compared to those characteristic of the \textsc{UN} dataset. 
%
%
Typically, (dynamic) topic models fitted on the \textsc{UN} dataset tend to infer topics which circle about e.g. war, peace or climate. In contrast, topic models trained on, say, \textsc{NeurIPS}, generically infer more varied topics, ranging from e.g. Neural Networks and their training to Reinforcement Learning.
See, for example, Table 6 in the supplementary material provided within the repository of our code \cite{source_code}, which shows six randomly sampled topics from each dataset as inferred by \texttt{NDF-TM}.

\begin{figure}[t!]
\begin{subfigure}{0.96\columnwidth}
  \centering
  \includegraphics[width=\linewidth]{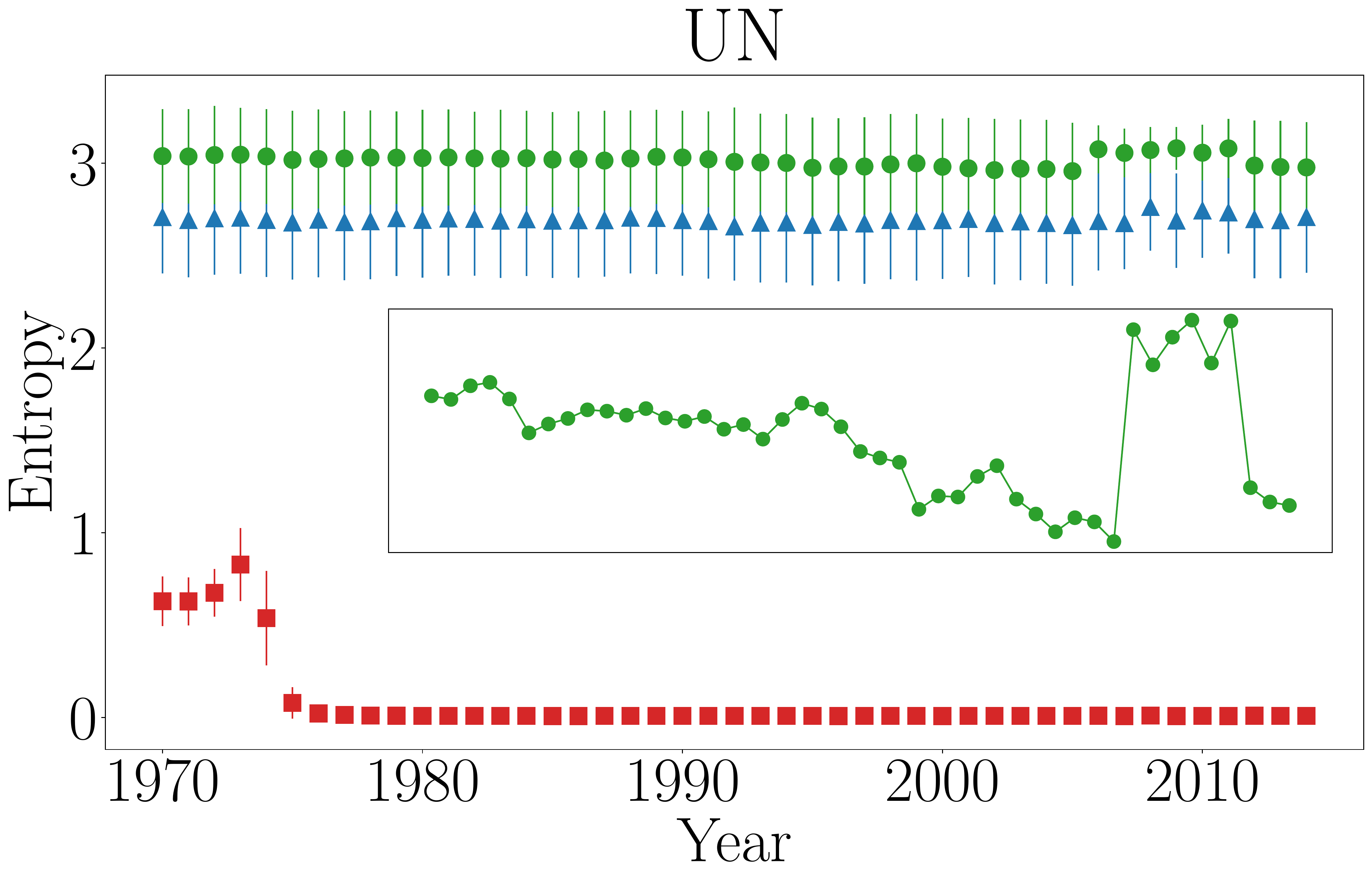} 
\end{subfigure}
\begin{subfigure}{0.98\columnwidth}
  \centering
  \includegraphics[width=\linewidth]{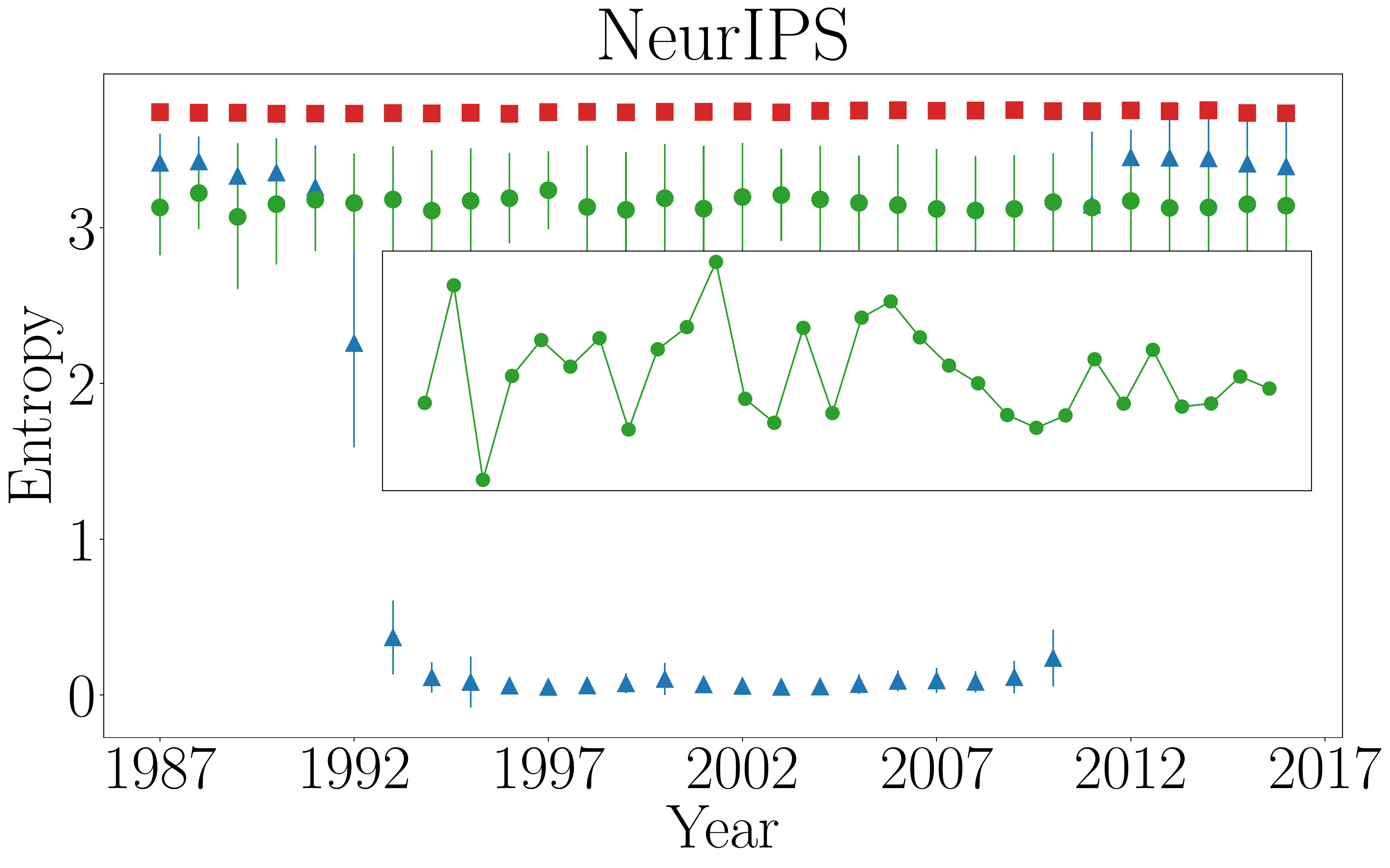}  
\end{subfigure}
\caption{Entropy of topic distribution inferred by \texttt{DTM-REP} (squares), \texttt{D-ETM} (triangles) and \texttt{NDF-TM} (circles), averaged over documents as time evolves. Values shown with one standard deviation for both \textsc{UN} (above) and \textsc{NeurIPS} (below) datasets. Note that the maximum entropy value is $\log(K=50) \approx 3.9$. The inset shows the details of the time-dependent topic-proportion entropy featured by \texttt{NDF-TM}. Note how the entropy decreases with time in the \textsc{UN} dataset (leaving aside the pick around the window 2005-2010) but fluctuates strongly for the (skewed) \textsc{NeurIPS} dataset.}
\label{fig:entropy_ts}
\end{figure}

It is easy to imagine that the more generic topics in the \textsc{UN} dataset (like war, climate, etc) have reached some type of equilibrium and thus display overall a less skewed distribution over the document collection. 
If this were the case, topic models featuring the proportion-activity coupling would fit well the data by only inferring the more generic topics. 
Figure \ref{fig:entropy_ts} shows the (Shannon) entropy of the topic distribution, averaged  over documents as time evolves, as inferred by all models.\footnote{The Shannon entropy of the topic distribution per document and time is defined here by $H_{t, d}=- \sum_i^K \theta^{(i)}_{t, d} \log \theta^{(i)}_{t, d}$, where $\theta^{(i)}_{t, d}$ is the $i$th component of $\+\theta_{t,d}$.} 
Note how the entropy inferred by \texttt{DTM-REP} (which features the proportion-activity coupling) for \textsc{UN} is close to zero, meaning that \texttt{DTM-REP} usually describes the documents with few topics, whereas for \textsc{NeurIPS} the entropy of the average topic distribution is close to its maximum value ($\log(K=50)\approx3.9$), meaning that it allocates almost equal probability for all $K$ topics (that is, the model needs all topics to fit the data well), as expected for a skew topic distributions. 
In contrast, \texttt{NDF-TM} uses the additional Bernoulli variable sequences to redistribute the noise in the topic dynamics. 
%
%
Note also how the topic entropy of \texttt{D-ETM} is often similar to that of \texttt{NDF-TM}, meaning \texttt{D-ETM} does in fact implicitly lift the proportion-activity coupling.

\begin{figure}[!t]
    \centering
    \includegraphics[width=\columnwidth]{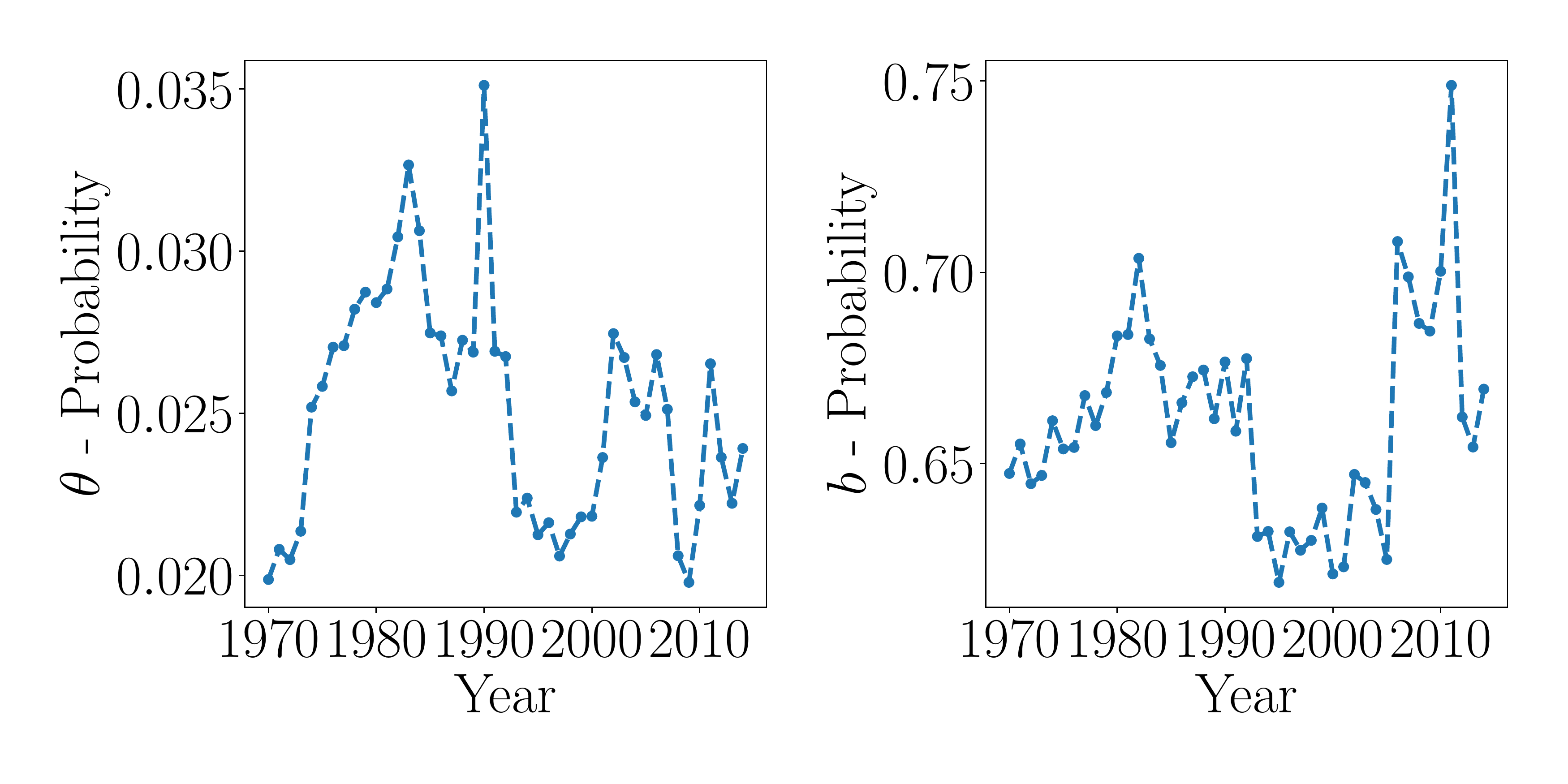}
    \caption{Evolution of topic proportion and activity probability for the topic \textit{middle east} inferred from the \textsc{UN} dataset via \texttt{NDF-TM}.}
    \label{fig:timeseries}
\end{figure}

Figure \ref{fig:timeseries} shows our results for one topic inferred from the \textsc{UN} dataset, namely \textit{middle east}. 
Note, for example, that the topic proportion for this topic peaks in the year 1990, which coincides with the Gulf War, to then drop right after.
Such a drop is also reflected in the topic activity. 
Later, in 2011, the Syrian Civil War started. 
This event is captured by the topic activity which peaks at 2011, even though the topic proportion probability is decreasing.
That is, even when the proportion of the \textit{middle east} topic is low within the documents of that year, it must remain active to properly describe the data.
%

\section{Conclusion}

We have introduced the Neural Dynamic Focused Topic Model for sequentially collected data, which explicitly decouples the dynamic topic proportions from the topic activities through the addition of sequences of Bernoulli variables.
We have shown that our approach consistently yields coherent and diverse topics, which correctly capture historical events.
Future work includes using \texttt{NDF-TM} together with Variational Autoencoders for topic-guided text generation.

\section{Acknowledgments}
This research has been funded by the Federal Ministry of Education and Research of Germany and the state of North-Rhine Westphalia as part of the Lamarr-Institute for Machine Learning and Artificial Intelligence, LAMARR22B.

C\'esar Ojeda is supported by Deutsche Forschungsgemeinschaft (DFG) - Project-ID 318763901 - SFB1294.

\bibliography{aaai23}

\end{document}